\documentclass[letterpaper]{article} 
\usepackage{aaai25}  
\usepackage{times}  
\usepackage{helvet}  
\usepackage{courier}  
\usepackage[hyphens]{url}  
\usepackage{graphicx} 
\usepackage{natbib}  
\usepackage{caption} 
\usepackage{amssymb}
\usepackage{algorithm}
\usepackage{algorithmic}
\usepackage{booktabs}
\usepackage{xcolor}
\usepackage{amsmath}

\usepackage{newfloat}
\usepackage{listings}
\usepackage{xspace}
\DeclareCaptionStyle{ruled}{labelfont=normalfont,labelsep=colon,strut=off} 
\floatstyle{ruled}
\newfloat{listing}{tb}{lst}{}
\floatname{listing}{Listing}
\newcommand{\folktables}{\texttt{folktables}\xspace}
\newcommand{\base}{\textsc{Base}\xspace}
\newcommand{\fairbase}{\textsc{FairBase}\xspace}
\newcommand{\mpmc}{FM$_{\text{MC}}$\xspace}
\newcommand{\mpma}{FM$_{\text{MA}}$\xspace}
\newcommand{\fmeo}{FM$_{\text{EO}}$\xspace}
\newcommand{\fmdp}{FM$_{\text{DP}}$\xspace}
\newcommand{\enforcema}{\textsc{Enforce}$_{\text{MA}}$\xspace}
\newcommand{\enforcemc}{\textsc{Enforce}$_{\text{MC}}$\xspace}

\newcommand{\mixup}{\textsc{Mixup}\xspace}
\newcommand{\mixupeo}{\textsc{Mixup}$_{\text{EO}}$\xspace}
\newcommand{\mixupma}{\textsc{Mixup}$_{\text{MA}}$\xspace}
\newcommand{\mixupmc}{\textsc{Mixup}$_{\text{MC}}$\xspace}
\newcommand{\mixupenforcemc}{\textsc{Mixup$_{\textsc{Enforce}_{\text{MC}}}$}\xspace}

\newcommand{\all}{\textsc{All}\xspace}
\newcommand{\bigger}{\textsc{Big}\xspace}
\newcommand{\smaller}{\textsc{Small}\xspace}
\newcommand{\dis}{\textsc{Dis}\xspace}
\newcommand{\dislfr}{\textsc{DLFR}\xspace}

\newcommand{\employment}{\textsc{Employment}\xspace}
\newcommand{\income}{\textsc{Income}\xspace}

\title{Who's the (Multi-)Fairest of Them \all: Rethinking Interpolation-Based Data Augmentation Through the Lens of Multicalibration}
\author{
    Karina Halevy,\textsuperscript{\rm 1 2}
    Karly Hou,\textsuperscript{\rm 2}
    Charumathi Badrinath\textsuperscript{\rm 2}
}
\affiliations{
    \textsuperscript{\rm 1}Carnegie Mellon University\\
    \textsuperscript{\rm 2}Harvard University\\
    khalevy@andrew.cmu.edu
}

\begin{document}
\maketitle

\begin{abstract}
Data augmentation methods, especially SoTA interpolation-based methods such as Fair Mixup, have been widely shown to increase model fairness. However, this fairness is evaluated on metrics that do not capture model uncertainty and on datasets with only one, relatively large, minority group. As a remedy, multicalibration has been introduced to measure fairness while accommodating uncertainty and accounting for multiple minority groups. However, existing methods of improving multicalibration involve reducing initial training data to create a holdout set for post-processing, which is not ideal when minority training data is already sparse. This paper uses multicalibration to more rigorously examine data augmentation for classification fairness. We stress-test four versions of Fair Mixup on two structured data classification problems with up to 81 marginalized groups, evaluating multicalibration violations and balanced accuracy. We find that on nearly every experiment, Fair Mixup \textit{worsens} baseline performance and fairness, but the simple vanilla Mixup \textit{outperforms} both Fair Mixup and the baseline, especially when calibrating on small groups. \textit{Combining} vanilla Mixup with multicalibration post-processing, which enforces multicalibration through post-processing on a holdout set, further increases fairness.
\end{abstract}

%
\begin{links}
    \link{Code}{https://github.com/ENSCMA2/fairest-mixup}
    \link{Proceedings version}{https://ojs.aaai.org/index.php/AAAI/article/view/33870}
\end{links}

\section{Introduction}

\begin{figure*}[th]
    \centering
    \includegraphics[width=0.9\linewidth,height=190px]{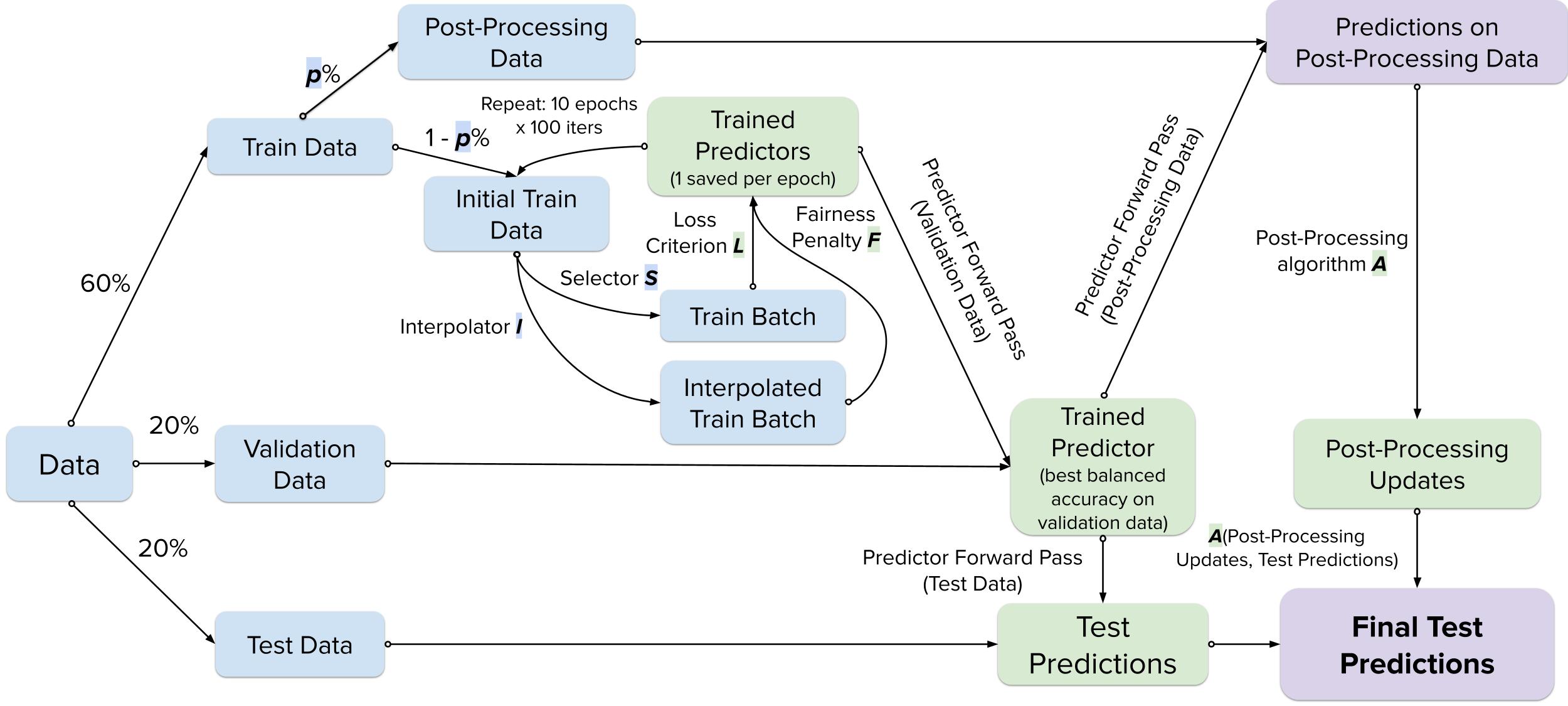}
    \caption{The ML training and evaluation pipelines considered in our work. Each method in our experiments can be characterized by a unique combination of: a percentage $p$ of post-processing data taken from training data, an interpolation-based data augmentation method $I$, a training batch selection procedure $S$, a training loss criterion $L$, a training fairness penalty $F$, and a post-processing algorithm $A$. These unique combinations are listed in Table \ref{tab:methods}.}
    \label{fig:fig1}
\end{figure*}

Algorithmic fairness has become increasingly important with the ubiquitous application of machine learning (ML). Unfairness can arise from many sources \cite{Huang2022-oy}, including unequal representation of protected groups in data \cite{Guo2022-vx}. For example, people of color can be underrepresented in clinical trials due to access barriers, lack of information, and discrimination \cite{Allison2022-nv}, leading ML models to have trouble predicting treatment outcomes for non-white patients. One way to mitigate underrepresentation is data augmentation, which creates synthetic individuals from the original data \cite{chuang2021fair,iosifidisdealing,chawla2002smote,sharma2020idealworld}. A particularly promising form of augmentation is Mixup \cite{DBLP:journals/corr/abs-1710-09412} and its fairness-oriented counterpart Fair Mixup \cite{chuang2021fair}, which linearly interpolate individuals with features in between majority and minority group attributes. However, existing augmentation literature measures fairness through binary metrics like demographic parity and equalized odds \cite{chuang2021fair}, which accumulate loss even when predictors lean toward correct labels. These metrics can be misleading because data often does not include all predictive features, so some notion of uncertainty is appropriate in a good predictor but would be penalized. Furthermore, the methods in \citet{chuang2021fair} only assess and optimize fairness for one minority group, but a fair predictor should work well on multiple multi-dimensional intersecting groups.

The metric of multicalibration (MC) \cite{pmlr-v80-hebert-johnson18a} accounts for this uncertainty and for the presence of multiple groups by comparing predicted probabilities to true probabilities, averaging over groups of interest, and considering subsets of a predictor's support separately. \citet{pmlr-v80-hebert-johnson18a} also introduce an algorithm, with runtime inversely proportional to the size of the smallest group, to post-process a predictor using a holdout set and guarantee a maximum MC violation. \citet{covid_paper} then use this algorithm to learn prediction adjustments from a holdout set and apply those adjustments to test predictions. However, such post-processing subtracts a substantial amount of holdout data from available training data, resulting in even less representation of underrepresented groups in initial training. Moreover, with runtime inversely proportional to group size, enforcing MC for very small groups can be slow. The guarantees of MC enforcement and upper bounds of the overall accuracy tradeoffs proven in \citet{pmlr-v80-hebert-johnson18a} also only apply to the post-processed holdout set, not to unseen test data.

This work examines whether we can combine the desirable properties of MC and data augmentation to supplement the binary outcome insights that demographic parity and equalized odds provide. We ask:
\begin{enumerate}
    \item RQ1: Under what conditions can Fair Mixup mitigate MC violations of neural network predictors on minority groups while preserving binary classification accuracy?
    \item RQ2: When can (Fair) Mixup serve as an alternative to and/or increase the efficiency of MC post-processing?
    \item RQ3: What aspects of Fair Mixup contribute to its success or failure in improving MC-based fairness?
\end{enumerate}
We contribute the first MC-based investigation of several (Fair) Mixup- and MC-inspired neural network training methods (depicted in Figure \ref{fig:fig1}), stress-testing performance and fairness on intersecting demographic groups and creating a new perspective on whether data augmentation is effective. We find that Fair Mixup can only mitigate MC violations and outperform post-processing under its original design of optimizing one group at a time. However, vanilla Mixup consistently makes predictors fairer and results in an average balanced accuracy/MC violation improvement of up to 14.22\% when combined with MC post-processing. We also find that the key performance-enhancing component of Fair Mixup is that it learns from interpolated data points. However, its other components (balancing training data by minority group membership and penalizing pairwise unfairness during training) detract from baseline performance, resulting in average balanced accuracy/MC violation decreases of up to 12.29\%.

\section{Preliminaries}
This section defines calibration \cite{pmlr-v80-hebert-johnson18a,chouldechova2017fair}, multicalibration \cite{pmlr-v80-hebert-johnson18a}, multiaccuracy \cite{pmlr-v80-hebert-johnson18a}, and  the data augmentation methods we later expand on. 
\subsection{Notation}
Throughout this paper, $\mathcal{X}$ represents a universe of individuals, $x_i$ represents an individual with index $i$, $S \subseteq \mathcal{X}$ is a subset of individuals, $\mathcal{C} \subseteq 2^{\mathcal{X}}$ is a set of subsets of individuals, $f$ is a predictor that maps individual $x_i$ to outcome probability $f_i$, $p_i^*$ is the true outcome probability of $x_i$, and $y_i \in \{0, 1\}$ is the binarized true outcome for $x_i$.
\subsection{Calibration}
For a maximum violation $\alpha \in [0,1]$, $f$ is $\alpha$\textbf{-calibrated} w.r.t. $S$ if $\exists S' \subseteq S$ with $|S'| \geq (1 - \alpha) |S|$ such that $\forall v \in [0,1],$
\begin{equation}\label{eq:calibration}
    |\mathbb{E}_{x_i \sim (S_v \cap S')}[f_i - p_i^*]| \leq \alpha,
\end{equation}
where $S_v = \{x_i:f_i = v\}$. In most classification tasks, we only see the binary outcome $y_i$ for $x_i$. Thus, we use a modification called \textbf{observable} calibration \cite{pmlr-v80-hebert-johnson18a}, where $y_i$ replaces $p_i^*$ in Eq. \ref{eq:calibration}.

For example, a tumor malignancy classifier is $0.05$-observably calibrated for $v = 0.6$ on Latine patients if of all Latine patients for which it predicts a 60\% chance of malignancy, 55\% to 65\% of these patients have a malignant tumor. The classifier is $0.05$-observably calibrated on Latine patients if this holds for all $v$---of all Latine patients for which it predicts a $v$ chance of malignancy, between $v - 5\%$ and $v + 5\%$ of these patients have a truly malignant tumor.

\subsection{Multicalibration}
$f$ is \textbf{$\mathbf{(\mathcal{C},\alpha)}$-multicalibrated} if it is $\alpha$-calibrated w.r.t. all $S \in \mathcal{C}$ \cite{pmlr-v80-hebert-johnson18a}. We define MC as in \citet{pmlr-v80-hebert-johnson18a}, but we require $S = S'$ (calibration on all of $S$ rather than any $1 - \alpha$ of it, explained in Appendix \ref{app:implementation}). For computational feasibility over datasets with millions of prediction probabilities, we also discretize the predicted probabilities. For integer $d > 0$, the $d$\textbf{-discretized} version of $S$ splits $S$ into $d + 1$ subsets, where
\begin{align}
    S_{v} = \{x_i: \frac{v}{d} \leq f_i < \frac{v + 1}{d}\} \text{ for } v \in [0,1,...,d].
\end{align}

Continuing with the tumor malignancy classifier example, the subset of the $10$-discretized $S = $ Latine patients with $v = 6$ would be all Latine patients with a predicted chance of at least $60\%$ but less than $70\%$ malignancy. Suppose all patients in this subset have a prediction of $63\%$. $0.05$-calibration would require that $58$ to $68\%$ of these patients have a truly malignant tumor, and that the corresponding conditions hold for all other $v \in [0, 10]$. Given $\mathcal{C} = \{$Black patients, Asian patients, Latine patients$\}$, $\mathcal{C}, 0.05$-multicalibration requires that this $0.05$-calibration must hold for Black, Asian, and Latine patients.
\subsection{Multiaccuracy}
Multiaccuracy (MA) \cite{pmlr-v80-hebert-johnson18a} is a looser version of MC. $f$ is $(\mathcal{C}, \alpha)$\textbf{-multiaccurate} if $ \forall S \in \mathcal{C}$,
\begin{equation}
    |\mathbb{E}_{x_i \sim S}[f_i - p_i^*]| \leq \alpha.
    \end{equation}
Rather than requiring the expected prediction error within each $S$ \textit{and} predicted probability to be $\leq \alpha$, MA only requires this error to be $\leq \alpha$ in $S$ overall. Thus, for $\mathcal{C}$ = $\{$Black patients, Latine patients$\}$, $(\mathcal{C}, 0.05)$-multiaccuracy means that the average prediction for Black patients is within 5\% of the true proportion of Black patients that have a malignant tumor, and likewise for Latine patients. In the rest of this paper, when we say $f$ has an MC or MA violation of $\alpha$ on $\mathcal{C}$, we mean that $\alpha$ is the smallest value for which $f$ is $(\mathcal{C}, \alpha)$-multicalibrated or multiaccurate.
\subsection{Mixup}\label{sec:mixup}
Mixup was proposed to improve the generalizability of neural networks (NN) by training on linear combinations of example pairs, with the intuition that the NN would learn how predictions differ as inputs move continuously between feature sets \cite{DBLP:journals/corr/abs-1710-09412}. For training batch size $b$, mixup draws $(x_1,y_1) ,...,(x_b, y_b)$ and $(x_1', y_1'),...,(x_b', y_b')$ without replacement from the training data. Let $t \sim \text{Beta}(\epsilon, \epsilon)$ where $\epsilon \in (0, \infty)$. Mixup constructs one synthetic point per $i \in [1, ..., b]$:
\begin{align}
    (x_i'',y_i'') = (t x_i + (1 - t)x_i', ty_i + (1 - t)y_i')
\end{align}
and trains an NN on $(x_1'', y_1''), \cdots, (x_b'', y_b'')$ instead of the original batch. \citet{DBLP:journals/corr/abs-1710-09412} showed that mixup decreased test error on CIFAR-10 and CIFAR-100.
\subsection{Fair Mixup}
\citet{chuang2021fair} adapted mixup toward the goal of fairness. \textbf{Fair Mixup} (FM) samples $(x_1,y_1) ,...,(x_b, y_b)$ from minority group $S$ and $(x_1', y_1'),...,(x_b', y_b')$ from $S' = \neg S$. Mixup is then performed on these samples as in Section \ref{sec:mixup} to create synthetic points. The loss function applies the standard Binary Cross Entropy (BCE) loss function to the original points, applies the gradient $\mathcal{R}_{\text{mixup}}^{\mathcal{M}_S}$ of a pairwise fairness penalty $\mathcal{M}$ between $S$ and $S'$ to the synthetic points, and adds $\lambda$ times the fairness penalty to the BCE. Fair Mixup creates better tradeoffs between average precision and the fairness metrics of demographic parity and equalized odds \cite{chuang2021fair}.

\section{Related Work}
\subsection{Data Augmentation for Fairness}
There are several other data augmentation methods for fairness. In oversampling, minority group samples are duplicated until equal in number to majority group samples \cite{iosifidisdealing}. Another method, SMOTE, creates minority group members through linear interpolation among existing minority group members \cite{chawla2002smote}. More recently, \citet{sharma2020idealworld} introduce ``Ideal World'': for each original point, a new sample is created with the same features and label, but the protected attribute is flipped, making both statistical parity difference and average odds difference decrease while preserving accuracy. Outside of structured data, \citet{Wadhwa} apply identity pair replacement, identity term blindness, and identity pair swap on text classification. \citet{dur30710} introduce data augmentation that improves facial recognition on minority groups.

We focus on structured data classification to minimize the confounding factor of unstructured data featurization. We also choose Fair Mixup as a basis because it minimizes data distribution changes and treats protected attributes as predictive features. Ideal World takes away the predictive information of protected attributes. Oversampling, SMOTE, and Ideal World create additional minority individuals, changing the frequency and composition of minority groups. In contrast, Fair Mixup creates individuals that are neither minority nor majority group members, but rather some interpolated in-between. Thus, while the data distribution may change, the members of boolean circuit-defined groups do not.

\subsection{Extensions of MC}
\citet{pmlr-v80-hebert-johnson18a} devise algorithms that could enforce MC $\alpha$'s to be below an arbitrary threshold. A related post-processing algorithm, designed for multiaccuracy, is \textsc{Multiaccuracy Boost}, which requires a trained auditor on top of a holdout set \cite{kim2019multiaccuracy}. Applying the results of \citet{pmlr-v80-hebert-johnson18a} empirically, \citet{covid_paper} transfer learned post-processing updates to a COVID-19 mortality rate forecasting task. We test this application in Section \ref{subsubsec:pp}. 

A few works extend (multi-)calibration to more nuanced metrics that handle complex notions of uncertainty. \citet{pmlr-v80-kumar18a} add calibration optimization to the training loss function, clamping overconfident predictions while minimizing penalties on true confident predictions. \citet{wald2021on} propose multi-domain calibration to evaluate model generalization to out-of-distribution data, suggesting both isotonic regression post-processing and a training regime that includes calibration from \citet{pmlr-v80-kumar18a}. \citet{pmlr-v134-jung21a} extend MC to higher moments, measuring moment consistency in a way that computes groupwise error inversely proportionally to group size \cite{pmlr-v134-jung21a}. Other work extends MC to conformal prediction, which generates prediction sets rather than point estimates \cite{jung2023batch,barber2020limits}. This framework generalizes MC to quantiles of the label's support rather than individual values and is useful for categorical or continuous labels, unlike binary labels, for which MC is already a probabilistic extension. \citet{omnipredictors-for-regression} connect MC to multi-group loss minimization.

The most comprehensive investigation of MC post-processing to our knowledge is \citet{hansen2024multicalibrationpostprocessingnecessary}, which finds that baseline predictors on tabular data are often decently multicalibrated already, and post-processing does not improve worst-group calibration error for multi-layer perceptrons, Random Forests, and Logistic Regression but does benefit Support Vector Machines, Decision Trees, and Naive Bayes. When worst-group calibration error improves, there is an overall accuracy tradeoff. They further find that MC enforcement is hyperparameter-sensitive and most effective with huge amounts of data (found in image and language data but not tabular data). They find that calibration algorithms like Platt scaling and isotonic regression sometimes perform nearly on par with MC enforcement while being more efficient. These findings are consistent with previous works suggesting that empirical risk minimization may inevitably yield multicalibrated baseline predictors \cite{blasiok2023when,blasiok2024loss}. We refer the reader to \citet{hansen2024multicalibrationpostprocessingnecessary} for a more comprehensive MC literature review and for image and language experiments.

This work extends \citet{hansen2024multicalibrationpostprocessingnecessary} in three ways. First, expanding upon their maximum of 15 groups that are all at least 0.5\% of their corresponding population, we stress-test our methods on MC w.r.t. up to 81 groups at a time, up to 55 of which are smaller than 0.25\% of their corresponding population. We also select these groups in five different ways to investigate effects of group set size on MC. Second, expanding upon their examination of income prediction from \texttt{folktables} on Californian residents from 2018, we evaluate our methods on each permutation of the 10 most populous US states and the four most recent American Community Survey data collection years, yielding 40 datasets. We additionally test employment status prediction on these 40 datasets, for a total of 80 tasks considered. Third, while their work and much of the current MC literature considers data-reductive post-processing methods, our work takes inspiration from their finding that post-processing works best on huge datasets and instead focuses on data augmentation to maximize the amount of original data that can be used for initial training.

\section{Methods}\label{sec:methods}
\begin{table*}[th]\fontsize{9pt}{9pt}\selectfont
    \centering
    \begin{tabular}{c|ccp{2.8cm}ccc}\toprule
      Method   &  $p$ & $I$ & $S$ & $L$ & $F$ & $A$  \\\midrule
       \base  & 0 & $I(\cdot) = []$ & uniform random & BCE & $F(\cdot) = 0$ & $A(\cdot) = []$\\
       \fairbase & 0 & $I(\cdot) = []$ & balance by group & BCE & $F(\cdot) = 0$ & $A(\cdot) = []$\\
       \mixup & 0 & Mixup & uniform random & $L(\cdot) = 0$ & BCE & $A(\cdot) = []$ \\
       \mixupeo & 0 & Mixup & balance by group $\times y_i$  & BCE & $\lambda \cdot \text{BCE}$ & $A(\cdot) = []$ \\
       \mixupma & 0 & Mixup & balance by group & BCE & $\lambda \cdot \text{BCE}$ & $A(\cdot) = []$ \\
       \mixupmc & 0 & Mixup & balance by group $\times f_i$ & BCE & $\lambda \cdot \text{BCE}$ & $A(\cdot) = []$ \\
       \fmdp & 0 & Mixup & balance by group  & BCE & $\lambda \cdot \mathcal{R}_{\text{mixup}}^{\text{DP}}$ & $A(\cdot) = []$ \\
       \fmeo & 0 & Mixup & balance by group $\times y_i$ & BCE & $\lambda \cdot \mathcal{R}_{\text{mixup}}^{\text{EO}}$ & $A(\cdot) = []$ \\
       \mpma & 0 & Mixup & balance by group & BCE & $\lambda \cdot \mathcal{R}_{\text{mixup}}^{\text{MA}}$ & $A(\cdot) = []$ \\
       \mpmc & 0 & Mixup & balance by group $\times f_i$ & BCE & $\lambda \cdot \mathcal{R}_{\text{mixup}}^{\text{MC}}$ & $A(\cdot) = []$ \\
       \enforcema & 25 & $I(\cdot) = []$ & uniform random & BCE & $F(\cdot) = 0$ & Listing \ref{lst:ma}\\
       \enforcemc & 25 & $I(\cdot) = []$ & uniform random & BCE & $F(\cdot) = 0$ & Listing \ref{lst:mc}\\
       \mixupenforcemc & 25 & Mixup & uniform random & $L(\cdot) = 0$ & BCE & Listing \ref{lst:mc} \\
         \bottomrule
    \end{tabular}
    \caption{Post-processing data split percentages $p$, data augmentors $I$, training batch selectors $S$, loss criteria $L$ (applied to original data), fairness penalties $F$ (applied to synthetic data), and post-processing algorithms $A$ that uniquely characterize each method described in Section \ref{sec:methods} and diagrammed in Fig. \ref{fig:fig1}. BCE stands for Binary Cross Entropy loss.}
    \label{tab:methods}
\end{table*}

\begin{figure}
    \centering
    \includegraphics[width=0.93\linewidth,height=140px]{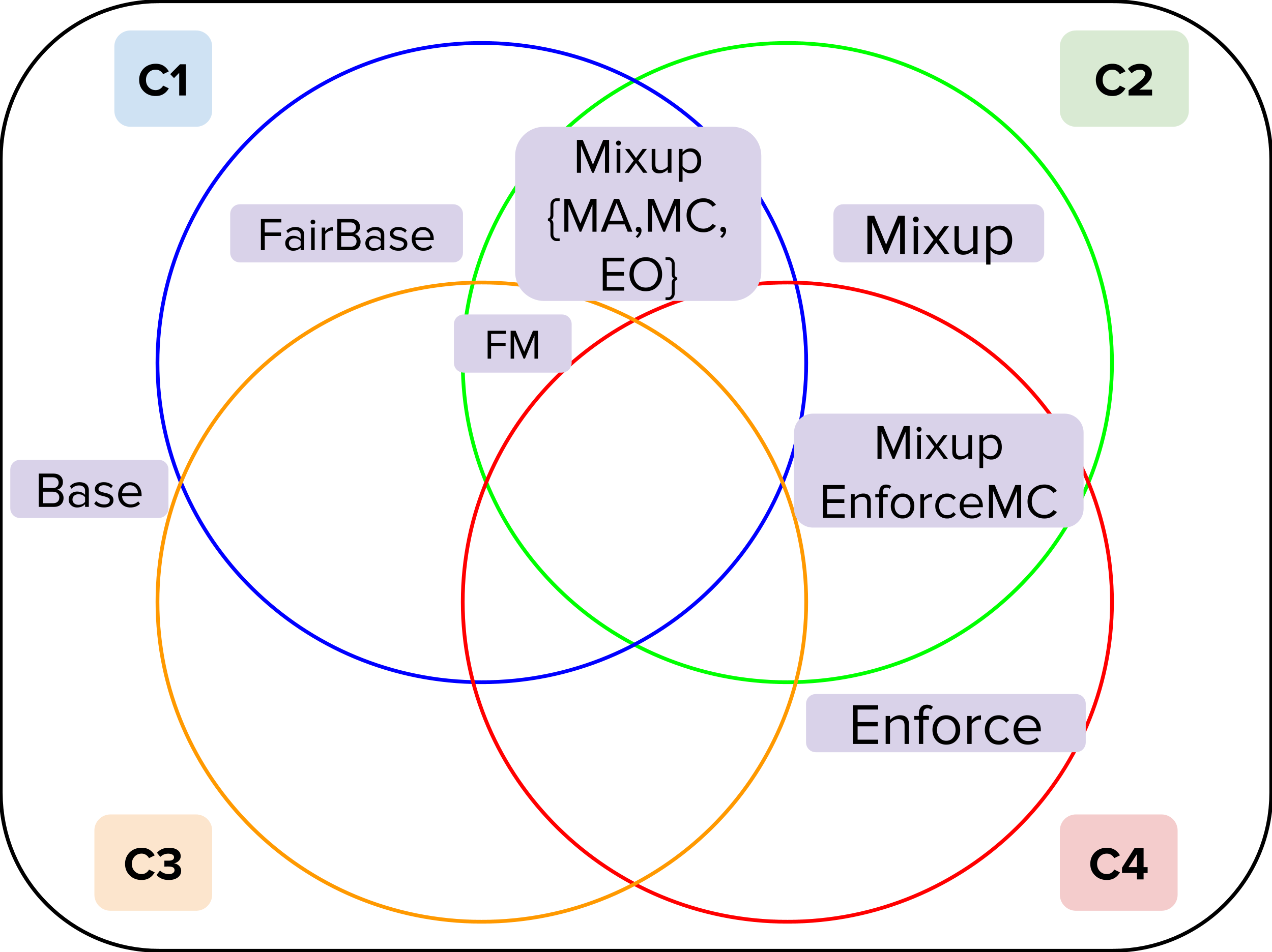}
    \caption{Venn Diagram of each method's core components.}
    \label{fig:methods}
\end{figure}
We test 13 NN training methods to determine the effects of particular features of FM that contribute to its performance and fairness. FM has 3 distinguishing components: \begin{enumerate}
    \item C1: Training batches are balanced across membership in the minority group for which we wish to ensure fairness.
    \item C2: Synthetic data is created by linearly interpolating original points. If C1 is implemented, each synthetic data point is the interpolation of a minority group member and a majority group member. If not, the original points are split in half and paired at random for interpolation.
    \item C3 (can only be done if C2 is also implemented): A fairness penalty is added to the loss function for predictions on synthetic points, minimizing a weighted sum of the standard loss and the fairness penalty during training.
\end{enumerate}
Post-processing is distinguished by the following: \begin{enumerate}\setcounter{enumi}{3}
    \item C4: A post-processing algorithm learns prediction update rules from post-processing data (subtracted from initial training data), and it applies those update rules to the validation and test data during evaluation and deployment.
\end{enumerate}
With these insights (summarized in Fig. \ref{fig:methods}), this section describes each method mathematically. We motivate each method by explaining how it implements a subset of $\{$C1, C2, C3, C4$\}$, thus isolating the effects of specific components of FM to answer RQ3. C4 also helps answer RQ2 (FM vs{.} post-processing). Method names are starred if they contain substantial novel elements that we introduce on top of existing work. Implementation details are in Appendix \ref{app:implementation}.
\subsection{Baselines}
\subsubsection{\base} trains an NN with mini-batch gradient descent using Binary Cross Entropy loss, over several epochs and batch selection iteration. We report test-time balanced accuracy for the epoch with the best validation-time balanced accuracy. \base does not implement C1, C2, C3, or C4.
\subsubsection{$^*$\fairbase} modifies \base by balancing training data groupwise (C1). Suppose we have minority groups $\mathcal{C}$ to optimize for fairness and $n$ iterations of gradient descent per training epoch in \base. Then, \fairbase conducts $n \cdot |\mathcal{C}|$ iterations of gradient descent. Each iteration centers around one $S \in \mathcal{C}$: we construct a batch by selecting one sub-batch from $S$ and one sub-batch from its complement $\neg S$. We subsample the larger sub-batch to be equal in size to the smaller sub-batch to ensure balance across membership in $S$.
\subsection{Variants of Mixup}
\subsubsection{\mixup} is as defined in Section \ref{sec:mixup}, implementing C2.
\subsubsection{$^*$\mixupeo} modifies \fairbase. Consider minority groups $\mathcal{C}$ and $n \cdot |\mathcal{C}|$ iterations of gradient descent as in \fairbase. \mixupeo conducts $2n \cdot |\mathcal{C}|$ iterations of gradient descent, each centered around one pair $(S, y) \in (\mathcal{C}, \{0, 1\})$. We construct a batch by selecting one sub-batch of members of $S_y$ (members of $S$ whose true label is $y$) and one sub-batch of members of $S'_y$ (members of $\neg S$ whose true label is $y$). Next, we perform mixup by pairing each member of $S_y$ with a member of $S'_y$ within the batch and interpolating each pair. Our loss is a weighted sum of Binary Cross Entropy applied to the original batch and the same loss applied to the interpolated points. \mixupeo implements C1, C2, and a control version of C3 (standard loss instead of pairwise fairness, but number of groups under consideration for this loss is adjustable, as elaborated on in Section \ref{subsubsec:fmmod}). Thus, we can compare it to \fairbase to isolate the effect of C2.
\subsubsection{$^*$\mixupma} creates one balanced batch per $S \in \mathcal{C}$, as in \fairbase, yielding $n \cdot |\mathcal{C}|$ gradient descent iterations. We interpolate each batch by pairing members of $S$ with members of $\neg S$ and adding $\lambda$ times the Binary Cross Entropy on the interpolated points. \mixupma also implements C1, C2, and a control version of C3, though C1 is slightly different than in \mixupeo, allowing us to compare variations of C1.
\subsubsection{$^*$\mixupmc} creates $d + 1$ batches per $S \in \mathcal{C}$ by creating $d$-discretized intervals of $f_i$'s, yielding $(d + 1) \cdot n \cdot |\mathcal{C}|$ gradient descent iterations. For each $S_v^d$ (members of $S$ with predicted probability in $[\frac{v}{d}, \frac{v + 1}{d})$), we construct a batch with half its points from members of $S_v$ and the other half from members of $(\neg S)_v^d$. 
We interpolate and calculate loss as in \mixupeo. \mixupmc also implements C1, C2, and a control version of C3, providing another way to compare the specifics of C1.

\subsection{Variants of Fair Mixup}\label{subsec:fm}
FM implements C1, C2, and C3. Though \citet{chuang2021fair} introduce two versions of FM (with $\mathcal{M}$ as demographic parity difference and equalized odds difference), their framework generalizes to any pairwise fairness metric. We first show how to modify FM to accommodate multiple minority groups simultaneously. Then, we define the two versions of FM from \citet{chuang2021fair}, followed by two versions with new metrics. Specifically, our extensions try to incorporate some notion of MC in the fairness penalty, as we aim to minimize MC violations. These methods test the effects of varying the metric in C3.
\subsubsection{Modifying (Fair) Mixup for Multiple Groups}\label{subsubsec:fmmod}
Consider metric $\mathcal{M}$, its group gradients $ \mathcal{R}_{\text{mixup}}^{\mathcal{M}_{S_1}},..., \mathcal{R}_{\text{mixup}}^{\mathcal{M}_{S_{|\mathcal{C}|}}}.$
For $k \in \{1,...,|\mathcal{C}|\}$, the penalty is the mean of the $k$ highest group gradients. We take means because preliminary experiments show that sums produce higher MC $\alpha$s. We make $k$ adjustable to prevent overfitting. The rest of the computation proceeds as in \citet{chuang2021fair}. Given $\mathcal{M}_S$, the first step is to transform it into an integral via the Fundamental Theorem of Calculus so it can be computed for interpolated data (Appendix A.1 in \citet{chuang2021fair}). Then, we differentiate the integral to get $\mathcal{R}_{\text{mixup}}^{\mathcal{M}_{S}}$. We list the equations for $\mathcal{M}_S$ below, with full formulae in Appendix \ref{app:fm}. 
\subsubsection{\fmdp} is the first version of FM in \citet{chuang2021fair}. $\mathcal{M}_S$ is demographic parity, the difference between the average $f_i$ on members of $S$ vs{.} its complement $S'$: \begin{equation}
    \mathcal{M}_S = \Delta \text{DP}_S(f) \\
    = |\mathbb{E}_{x_i \sim S}[f_i] - \mathbb{E}_{x_i \sim S'}[f_i]|.
\end{equation} 
\subsubsection{\fmeo} is the second version of FM, with the equalized odds difference \cite{hardt2016equality} that modifies DP by only considering one true outcome at a time: \begin{equation}
    \mathcal{M}_S = \Delta \text{EO}_{S}(f) \\
    = \sum_{y \in \{0, 1\}}|\mathbb{E}_{x_i \sim S_y}[f_i] - \mathbb{E}_{x_i \sim S'_y}[f_i]|,
\end{equation} where $S_y = \{x_i \in S: y_i = y\}$, and $S' = \neg S$. 
\subsubsection{$^*$\mpma} is our first extension of FM, with a version of MA modified to be pairwise. We measure the mean difference in prediction errors $e_i = f_i - p_i^*$ between $S$ and $S'$:
\begin{equation}
    \mathcal{M}_S = \Delta \text{MA}_S(f) \\
    = |\mathbb{E}_{x_i \sim S}[e_i] - \mathbb{E}_{x_i \sim S'}[e_i]|,
\end{equation} 
\subsubsection{$^*$\mpmc} is our second extension of FM, with a pairwise modification of MC, which modifies MA by considering one interval $S_{v}^d = \{ x_i \in S: f_i \in [\frac{v}{d}, \frac{v + 1}{d}) \}$ at a time:
\begin{equation}
    \mathcal{M}_S = \Delta \text{MC}_S(f) = \sum_{v = 0}^d |\mathbb{E}_{x_i \sim S_{v}^d}[e_i] - \mathbb{E}_{x_i \sim S'^d_{v}}[e_i]|.
\end{equation} 

\subsection{Post-Processing}\label{subsubsec:pp}
We test whether MC and MA enforcement (implementing C4) improve test performance as in \citet{covid_paper}.
\subsubsection{\enforcema} post-processes predictions to minimize MA violations. We feed (1) predictions on a holdout post-processing set and (2) a set of minority groups $\mathcal{C}$ as inputs to Algorithm 3.1 in \citet{pmlr-v80-hebert-johnson18a}. However, we augment Algorithm 3.1 with a list of rules mapping each $S \in \mathcal{C}$ to a float $a_S$ to be added to predictions on members of $S$. In other words, the algorithm learns how much to adjust predictions for each group. At validation and test time, we add $a_S$ to initial predictor outputs for members $S$.
\subsubsection{\enforcemc} post-processes predictions to minimize MC violations. It proceeds as in \enforcema, but we add an integer $d$ as a third input to Algorithm 3.2 in \citet{pmlr-v80-hebert-johnson18a}. We augment Algorithm 3.2 with a list of rules mapping each group $S_v$ (members of $S \in \mathcal{C}$ where $f_i \in [\frac{v}{d}, \frac{v + 1}{d})$), to a float $a_{S, v}$ to be added to predictions on members of $S$ whose initial predictions are in $[\frac{v}{d}, \frac{v + 1}{d})$.
\subsubsection{\mixupenforcemc} performs \mixup on a reduced initial training set followed by \enforcemc on a holdout post-processing set. This method tests C2 $\cup$ C4, as we ultimately find that \mixup performs best overall among methods that do not implement C4. We implement \mixupenforcemc to answer the part of RQ2 that asks whether data augmentation can improve the performance of \enforcemc.

\section{Experiments}
This section describes our data and experimental settings.

\subsection{Datasets}
\begin{table*}[ht]\fontsize{9pt}{9pt}\selectfont
    \centering
    \begin{tabular}{ccp{2.3cm}ccp{1.4cm}p{1.8cm}p{1.75cm}}\toprule
     Dataset & Size & \# Features (Binary, Categorical, Continuous) & \# Non-White  & \# Disabled & Max \# Minority Groups  & Max \# Groups $\leq$ 0.25\% of Population & Mean Size of Smallest Group \\\midrule
    \employment & 6,993,839 & 5, 9, 2 & 2,160,161 & 1,036,251 & 81 & 55 & 28.5 \\
    \income    & 3,543,292 & 2, 3, 6 & 1,014,632 & 250,074 & 51 & 28 & 31.85\\
         \bottomrule
    \end{tabular}
     \caption{Summary statistics of the \employment and \income datasets. ``Size'' is the number of individuals summed over all 40 subsets. Maxes and means are taken over these subsets. ``Smallest Group'' disabled members of the LFR.}
    \label{tab:data_stats}
\end{table*}
We test two prediction tasks from \folktables \cite{ding2021retiring}, a superset of Adult Income data \cite{KohaviBecker} collected from the American Community Survey. We have $p_i^* \in \{0, 1\}$, but $f_i \in [0, 1]$. Table \ref{tab:data_stats} summarizes the data. Full data statistics are at \url{http://tiny.cc/mfm-stats}.

\paragraph{\employment} The task is to predict whether an individual is employed. Table \ref{tab:employment_features} specifies the exact input features.

\paragraph{\income} The task is to predict whether an individual's annual income is higher than the median income for that year in their state of residence according to Data Commons \cite{datacommons}. Table \ref{tab:income_features} lists input features.

We run 40 datasets each for \employment and \income: the 10 most populous US states $\times$ the 4 most recent years, providing substantial geographic and temporal variation. We choose these tasks based on experiments in \citet{jung2023batch}. Additionally, we seek problems with reasonable baseline performance ($\geq 80\%$ balanced accuracy on CA $\times$ 2022) to focus on improving fairness on useful classifiers.
\subsection{Experimental Settings}
To measure the effects of $|\mathcal{C}|$ and $|S|$, we run all combinations of datasets and training methods on five settings:

\subsubsection{\all} $\mathcal{C} = \cup$ $\{$all $n$ computationally possible racial groups, disabled people, disabled members of each racial group$\}$. A racial group is computationally possible if for all random seeds, at least one disabled member of that group is in each of the train, validation, and test splits. $|\mathcal{C}| = \mathbf{2n + 1}$.
\subsubsection{\bigger} $\mathcal{C} = \cup$ \{$b$ racial groups each comprising $>$ 0.25\% of the total dataset, disabled people, disabled members of each of the $b$ racial groups.$\}$ $|\mathcal{C}| = \mathbf{2b + 1}, \mathbf{b << n}$.
\subsubsection{\smaller} $\mathcal{C} = \cup$ $\{s$ racial groups each comprising $\leq$ 0.25\% of the total dataset, disabled people, disabled members of each of the $s$ racial groups$\}$. $|\mathcal{C}| = \mathbf{2s + 1}, \mathbf{s << n}$.
\subsubsection{\dis} This setting is closest to what FM has already been tested on: $\mathcal{C} = \{\text{disabled individuals}\}$, so $|\mathcal{C}| = \mathbf{1}$.
\subsubsection{\dislfr} $\mathcal{C} = $ disabled people, members of the least frequent (computationally possible) racial group (LFR), and disabled members of the LFR, hence $|\mathcal{C}| = \mathbf{3}$.

\section{Results}
\begin{table*}[ht]\fontsize{9pt}{9pt}\selectfont
\setlength{\tabcolsep}{1mm}
\centering
  \begin{tabular}{p{2.5cm}|ccccc|ccccc}\toprule
Method &  & &\employment & & &  &  & \income &  \\
 &  \all & \bigger & \smaller & \dis & \dislfr & \all & \bigger & \smaller & \dis & \dislfr \\\midrule
 \base &	0.00	&	0.00	&	0.00	&	0.00	&	0.00	&	0.00	&	0.00	&	0.00	&	0.00	&	0.00 \\
    \fairbase & -2.69	&	-2.36	&	-4.28	&	2.08	&	-3.76	&	-3.04	&	-4.26	&	-11.08	&	-12.29	&	-6.34 \\
     \mixup &	2.89	&	3.11	&	2.56	&	-23.20	&	2.03	&	1.22	&	1.30	&	\textbf{3.12}	&	\textbf{3.20}	&	1.63	 \\
   \mixupeo &	-2.54	&	-2.21	&	-4.66	&	4.45	&	-4.07	&	-3.39	&	-4.14	&	-9.73	&	-12.09	&	-6.15 \\
   \mixupma &	-3.31	&	-3.08	&	-4.70	&	\textbf{4.50}	&	-5.35	&	-3.10	&	-3.51	&	-11.51	&	-10.80	&	-7.27 \\
    \mixupmc &	-2.91	&	-2.55	&	-5.30	&	3.09	&	-5.93	&	-3.08	&	-3.02	&	-10.34	&	-10.78	&	-6.58 \\ 
    \fmdp &	-3.41	&	-2.22	&	-5.23	&	3.84	&	-5.29	&	-3.34	&	-3.39	&	-10.75	&	-10.11	&	-7.67 \\
    \fmeo &	-2.26	&	-3.36	&	-4.05	&	1.73	&	-4.52	&	-3.55	&	-3.58	&	-10.00	&	-11.12	&	-5.55 \\
    \mpma &	-2.95	&	-3.99	&	-6.35	&	1.40	&	-5.79	&	-3.74	&	-3.41	&	-10.33	&	-11.92	&	-7.36 \\
    \mpmc &	-2.72	&	-2.36	&	-4.94	&	0.61	&	-4.12	&	-4.32	&	-3.23	&	-10.41	&	-11.30	&	-5.52 \\
     \enforcema &	-0.46	&	0.02	&	0.46	&	0.43	&	-0.32	&	-0.02	&	0.70	&	-1.27	&	-2.76	&	-1.66 \\
    \enforcemc  &	8.31	&	12.03	&	7.89	&	-15.28	&	\textbf{8.74}	&	9.87	&	9.97	&	-6.61	&	-17.82	&	8.28 \\
    \mixupenforcemc 	&	\textbf{10.36}	&	\textbf{12.97}	&	\textbf{9.23}	&	-29.27	&	8.72	&	\textbf{11.64}	&	\textbf{14.22}	&	-2.15	&	-11.37	&	\textbf{13.06} \\
    \bottomrule
\end{tabular}
\caption{Summary (mean of 10 trials) of methods across 40 (state, year) pairs $\times$ \employment and \income. Each number is the mean of the \% increase in balanced accuracy and \% decrease in worst-group MC $\alpha$.}
\label{tab:result_summary}
\end{table*}

To capture both fairness and overall performance, we compute the mean across all 40 (state, year) pairs of the following quantities for each experiment: (1) \% increase in balanced accuracy over \base for the corresponding state, year, and task and (2) \% decrease over \base in worst (highest) individual group MC violation $\alpha$. 
Table \ref{tab:result_summary} reports these mean percentages, showing that for all (task, setting) pairs except for \dis (both tasks), (\employment, \dislfr), and (\income, \smaller), \mixupenforcemc shows the biggest average balanced accuracy and MC $\alpha$ improvement. 
The other best methods are \mixupma for (\employment, \dis), \mixup for (\income, \dis) and (\income, \smaller), \enforcemc for (\employment, \dislfr) (though \mixupenforcemc is close), but \fmdp has the best $\alpha$ for (\employment, \dis) according to Table \ref{tab:worst_mc_summary}. If we consider only methods that perform post-processing or augmentation/data balancing (i.e. all methods except \mixupenforcemc), the best method is \enforcemc, except (\employment, \dis) (\mixupma was best), (\income, \smaller), and (\income, \dis) (\mixup was best). One note is that for 28 of 40 datasets, we had $s = 0$ and thus $\mathcal{C} = $ just disabled people, so (\income, \smaller) results may be more characteristic of single-group calibration. We also note that all methods except for \base had negative mean increases in balanced accuracy (up to -1.25\%), so positive values in Table \ref{tab:result_summary} indicate fairness improvements.

Examining FM, we see that except (\employment, \dis), all FM variants worsened fairness. For (\employment, \dis), FM improved fairness while largely preserving balanced accuracy, confirming the result in \citet{chuang2021fair} that FM works on one larger group. 

Comparing \mixupenforcemc and \enforcemc, we observe that while \mixupenforcemc outperforms \enforcemc in many cases, it sometimes makes the \enforcemc component of \mixupenforcemc less efficient. On the \employment dataset, the number of iterations to convergence of the \enforcemc post-processing algorithm increased by a percentage in the range $(0.34\%, 5.8\%)$, with the greatest percentage increase for \smaller (+5.8\%) and the greatest decrease for \bigger (-1.58\%). For \income, all methods took fewer iterations, in the range $(-3.68\%, -0.08\%)$. 

Finally, we analyze correlations between results and data statistics. We largely find either no correlation or low correlations, with some exceptions. One exception is that the mean MC $\alpha$ across groups $> 0.25\%$ of the population on \all has a moderate correlation with total dataset size (lower violations for bigger datasets) for all non-\base methods that do not involve \enforcemc. This mean $\alpha$ on \all also moderately correlates with the number of groups bigger than $0.25\%$ of the population for \mpma, and it also moderately correlates with number of groups smaller than $0.25\%$, total number of minority groups, number of disabled individuals, and number of non-white individuals for several methods. Looking at efficiency, the number of iterations to convergence of \enforcemc and \mixupenforcemc both strongly correlate with dataset size, but there is no correlation with the \% change in number of iterations between methods.

\section{Discussion}
Our results reveal the importance of stress-testing fairness optimization on multiple groups of varying sizes and on metrics that capture uncertainty. To answer RQ1, \textbf{the only condition under which FM improves MC is the condition it was designed for}: fairness for one minority group (\dis) on a truly binary problem (\employment). This holds irrespective of the particular train-time fairness penalty. This leads to an answer to RQ2: under a \textbf{single-group truly-binary condition}, FM (especially \fmdp) outperforms post-processing in ensuring fairness for disabled people. Based on the raw $\alpha$s in Table \ref{tab:worst_mc_summary}, this could be because disabled people are a relatively big, non-monolithic group for which the \base NN is already much more calibrated than the more fine-grained racial groups. Thus, to further improve upon the \base $\alpha$, it may be more effective to examine more disabled individuals and their full feature sets during training (as in FM) rather than apply a fixed adjustment to disabled individuals unconditionally (as in post-processing). However, this fixed post-processing adjustment may work well for smaller racial groups because the smaller sizes of racial groups make race more informative than disability.

\textbf{Regular \mixup presents a robust alternative} to \enforcema in nearly all settings and to \enforcemc when considering one group in a continuous-to-binary prediction problem as in (\income, \dis) or part of (\income, \smaller). More powerfully, \textbf{combining \mixup and \enforcemc through \mixupenforcemc enhances performance} of \enforcemc alone in the majority of settings, especially when more than one group is under consideration. However, it is inconclusive whether this enhancement is accompanied by efficiency improvement for the \enforcemc component.

For RQ3, we see that \textbf{\mixup is the overall best post-processing-free method}. Comparing \mixup with \mixupeo, \mixupma, \mixupmc, and \fairbase, we observe that \textbf{using interpolated data contributes more to fairness improvements} than groupwise balancing of training batches. Looking at \fairbase, \mixupeo, \mixupma, and \mixupmc, we further suggest that \textbf{data balancing may adversely affect performance and fairness}, since the key factor that sets \mixup apart from worse-performing methods of FM, \mixupeo, \mixupma, and \mixupma is C1. This may be because having limited minority instances means we learn less about majority instances as well (and since groups intersect, some instances that are minorities in one way but majorities in another are seen less). Finally, comparing \mixupeo, \mixupma, and \mixupmc to FM variants, we see that C3 effects (train-time fairness penalty) are inconclusive, as outcomes fluctuate by method and setting. Thinking more generally about why \mixup outperforms FM so often, we hypothesize that in addition to the adverse effect of data balancing in FM, \mixup has a more manageable amount of learning (normal BCE loss, with more data to learn from, net positive), while the pairwise fairness component of FM loss may be differently valued across demographic groups, thus possibly leading to less stable/effective learning (added complexity to the loss might also be a negative that worsens with the number of groups). 

\section{Conclusion}
We conduct the first investigation of how data augmentation via interpolation affects MC-based fairness on multiple minority groups of multiple sizes for binary tabular data classification. We find that while Fair Mixup is not so fair on multiple groups, regular mixup mitigates MC violations across many groups, both by itself and together with MC post-processing. Our investigation opens several avenues of future work, with our evaluation pipeline being easily extensible to data augmentation on probabilistic fairness in other modalities (e.g. vision, language) and ML problems (e.g. continuous/categorical labels).

\section*{Ethical Statement}\label{sec:ethics}
Augmentation introduces synthetic data and alters demographic representation to present the illusion that certain groups are well-represented. We urge creators and users of augmented datasets to be transparent about augmentation methods used. We lead by example as we release our datasets with full methodological descriptions. Furthermore, we caution that our implemented augmentation methods can substantially alter outcomes in real-world decision-making settings, and examining multicalibration is meant to be a supplement to, not a replacement for, frameworks addressing binary, individual-level fairness.

\section*{Acknowledgements}
We thank Professor Cynthia Dwork and Pranay Tankala for teaching the course that inspired this work and for mentoring us through the initial stages of the project. We also thank Professor Maarten Sap and Alfredo Gomez for helpful feedback during the paper writing and rebuttal process.

\appendix
\section{Dataset Details}\label{app:problemspec}
Table \ref{tab:employment_features} summarizes the input features of the \employment dataset, and Table \ref{tab:income_features} summarizes the features for \income. 
\begin{table}[th]\fontsize{9pt}{9pt}\selectfont
    \centering
    \begin{tabular}{p{2cm}p{1.6cm}cc}\toprule
       Feature  & Code & Type & Scale \\\midrule
       Detailed race code  & RAC3P & Categorical & 100 \\
       Relationship status  & RELP (2018) & Categorical & 18 \\
       Relationship status  & RELSHIPP (2019-2022) & Categorical & 19 \\
       Mobility status  & MIG & Categorical & 4 \\
       Military service   & MIL & Categorical & 5\\
       Ancestry  & ANC & Categorical & 4 \\
       Employment status of parents  & ESP &  Categorical & 9 \\
       Citizenship status  & CIT & Categorical & 5 \\
       Marital status  & MAR & Categorical & 5 \\
       Cognitive difficulty  & DREM &  Categorical & 3 \\
       Age  & AGEP & Continuous & 0-99 \\
       Educational attainment  & SCHL & Continuous & 0-24 \\
       Sex  & SEX & Binary & - \\
       Hearing difficulty  & DEAR & Binary & - \\
       Vision difficulty  & DEYE & Binary & - \\
       Born in US  & NATIVITY & Binary & - \\
       Disability  & DIS & Binary & - \\
         \bottomrule
    \end{tabular}
    \caption{Information about input features selected for \employment. Note on Scale column: for categorical variables, the Scale indicates the number of categories for the variable. For continuous variables, the Scale indicates the minimum and maximum of the range for the variable.}
    \label{tab:employment_features}
\end{table}
\begin{table}[th]\fontsize{9pt}{9pt}\selectfont
    \centering
    \begin{tabular}{p{2cm}p{1.8cm}cc}\toprule
       Feature  & Code & Type & Scale \\\midrule
       Detailed race code  & RAC3P & Categorical & 100 \\
       Class of worker  &  COW & Categorical & 10 \\
       Marital status  & MAR & Categorical & 5 \\
	Occupation	&   OCCP & Categorical & 531 \\
       Relationship status  & RELP (2018) & Categorical & 18 \\
       Relationship status  & RELSHIPP (2019-2022) & Categorical & 19 \\
	Place of birth & POBP & Categorical & 223 \\
	Age  & AGEP & Continuous & 0 - 99 \\
	Educational attainment  & SCHL & Continuous & 0 - 24 \\
		Usual hours worked per week in last 12 months &      WKHP & Continuous & 1 - 98 \\
	Sex  & SEX & Binary & - \\
	Disability  & DIS & Binary & - \\
         \bottomrule
    \end{tabular}
    \caption{Information about input features selected for our \income datasets from \folktables. Note on Scale column: for categorical variables, the Scale indicates the number of categories for the variable. For continuous variables, the Scale indicates the minimum and maximum of the range for the variable.}
    \label{tab:income_features}
\end{table}
\subsection{Hyperparameter Search}
We explore a few choices of $d$, $k$, and $\lambda$ for our \mixup and Fair Mixup implementations. For Fair Mixup, we try each of the 24 combinations of $d \in \{10, 55, 100\}$, $k \in \{1, 3, 40, 100\}$, and $\lambda \in \{0.25, 0.5\}$ on the subsets of each prediction task from California from the year 2022 (while running these combinations on all 40 subsets would be ideal, this one search took us over a week and thus would be intractable to replicate). We determine the best $(d, k, \lambda)$ triple for each (dataset, method) combination and use that triple on the other 39 (state, year) subsets for that dataset and method. We measure ``best'' via the highest average of (1) percent increase in balanced accuracy over \base and (2) percent decrease in mean individual-group MC violation over \base. 

For \mixup, we fix $d = 10$, state = CA, and year = 2022 and search over the 8 combinations of $k \in \{1, 3, 40, 100\}$ and $\lambda \in \{0.25, 0.5\}$, as $d = 55$ and $d = 100$ always dramatically worsened both efficiency and performance in our hyperparameter search for Fair Mixup. We measure and determine the best $(k, \lambda)$ for each (dataset, method) tuple as we do in our Fair Mixup search, and we apply these hyperparameters across all states and years within each dataset and method. 

One final hyperparameter is determining whether to create batches by sampling without replacement or by sampling the smaller group with replacement until it is equal to the specified batch size. We experiment with both options on California $\times$ 2022 $\times$ \all for all FM methods and determine that the best performance and fairness results from the following choices: (1) if the group is smaller than $183617.4$ (approximately 0.25\% of the \employment dataset size, multiplied by the training split percentage) \textit{and} the fairness metric is demographic parity or multiaccuracy, then we take a sample of size $b$, with replacement if $b$ is bigger than the size of the group, without replacement otherwise; (2) otherwise, we take a sample of size min(group size, $b$), without replacement.

Our final hyperparameter selections are in Table \ref{tab:hparams}.
                   
\begin{table}[!th]
    \centering
    \begin{tabular}{cc|ccc}\toprule
      Dataset & Method & $d$ & $k$ & $\lambda$ \\\midrule
       \employment  & \mixup & 10 & 3 & 0.25 \\
       & \mixupeo & 10 & 100 & 0.25 \\
       & \mixupma & 10 & 3 & 0.25 \\
       & \mixupmc & 10 & 40 & 0.25 \\
       & \fmdp & 10 & 100 & 0.5 \\
       & \fmeo & 10 & 100 & 0.25 \\
       & \mpma & 10 & 100 & 0.25 \\
       & \mpmc & 10 & 100 & 0.5 \\\hline
       \income & \mixup & 10 & 40 & 0.25 \\
       & \mixupeo & 10 & 40 & 0.5 \\
       & \mixupma & 10 & 40 & 0.25 \\
       & \mixupmc & 10 & 40 & 0.5 \\
       & \fmdp & 10 & 3 & 0.25 \\
       & \fmeo & 10 & 3 & 0.5 \\
       & \mpma & 10 & 3 & 0.5 \\
       & \mpmc & 10 & 3 & 0.25 \\ \bottomrule
    \end{tabular}
    \caption{Hyperparameters $d$, $k$, and $\lambda$ used for each (Fair) Mixup method and dataset.}
    \label{tab:hparams}
\end{table}
\section{Additional Results}\label{app:results}
Table \ref{tab:bacc_summary} gives the raw balanced accuracy percentages that contribute to the percent changes shown in Table \ref{tab:result_summary}, while Table \ref{tab:worst_mc_summary} gives the raw worst-group MC $\alpha$s that contribute to those same percent changes. Our full suite of summary statistics of results can be found at \url{http://tiny.cc/mfm-results}. Each number in the spreadsheet represents a mean over ten trials (random seeds 0 through 9).

\begin{table*}[ht]\fontsize{9pt}{9pt}\selectfont
\setlength{\tabcolsep}{1mm}
\centering
  \begin{tabular}{p{2.5cm}|ccccc|ccccc}\toprule
Method &  & &\employment & & &  &  & \income &  \\
 &  \all & \bigger & \smaller & \dis & \dislfr & \all & \bigger & \smaller & \dis & \dislfr \\\midrule
 \base &	\textbf{82.37}	&	82.37	&	\textbf{82.37}	&	82.37	&	\textbf{82.37}	&	\textbf{79.82}	&	\textbf{79.82}	&	\textbf{79.82}	&	\textbf{79.82}	&	\textbf{79.82} \\
    \fairbase &	81.96	&	82.05	&	82.01	&	82.33	&	82.19	&	78.94	&	78.99	&	79.04	&	78.93	&	79.08 \\
     \mixup &	81.64	&	81.64	&	81.64	&	81.64	&	81.64	&	79.67	&	79.67	&	79.67	&	79.67	&	79.67	 \\
   \mixupeo &	81.96	&	82.05	&	81.87	&	82.30	&	82.19	&	78.95	&	78.95	&	78.96	&	78.87	&	79.06 \\
   \mixupma &	81.95	&	82.04	&	81.93	&	82.27	&	82.07	&	78.90	&	78.90	&	78.93	&	78.91	&	78.93 \\
    \mixupmc &	81.99	&	82.06	&	81.93	&	82.29	&	82.05	&	78.89	&	78.91	&	78.92	&	78.94	&	78.88 \\ 
    \fmdp &	81.98	&	82.05	&	81.95	&	82.27	&	82.09	&	78.86	&	78.94	&	78.93	&	78.90	&	79.04 \\
    \fmeo &	81.96	&	82.04	&	81.89	&	82.31	&	82.18	&	78.93	&	78.97	&	78.95	&	78.93	&	79.10 \\
    \mpma &	81.97	&	82.06	&	81.91	&	82.29	&	82.16	&	78.90	&	78.92	&	78.98	&	78.92	&	78.91 \\
    \mpmc &	81.97	&	82.05	&	81.81	&	82.28	&	82.05	&	78.90	&	78.91	&	78.91	&	78.90	&	78.94 \\
     \enforcema &	82.35	&	\textbf{82.39}	&	82.31	&	\textbf{82.39}	&	82.32	&	79.60	&	79.67	&	79.64	&	79.65	&	79.52 \\
    \enforcemc  &	82.14	&	82.18	&	82.25	&	82.33	&	82.28	&	79.30	&	79.27	&	79.61	&	79.64	&	79.46 \\
    \mixupenforcemc 	&	81.57	&	81.65	&	81.51	&	81.77	&	81.64	&	79.22	&	79.27	&	79.56	&	79.58	&	79.45 \\
    \bottomrule
\end{tabular}
\caption{Balanced accuracies (\%, mean of 10 trials) of methods across 40 (state, year) pairs $\times$ \employment and \income.}
\label{tab:bacc_summary}
\end{table*}

\begin{table*}[ht]\fontsize{9pt}{9pt}\selectfont
\setlength{\tabcolsep}{1mm}
\centering
  \begin{tabular}{p{2.5cm}|ccccc|ccccc}\toprule
Method &  & &\employment & & &  &  & \income &  \\
 &  \all & \bigger & \smaller & \dis & \dislfr & \all & \bigger & \smaller & \dis & \dislfr \\\midrule
 \base &	0.674	&	0.639	&	0.660	&	0.121	&	0.609	&	0.668	&	0.666	&	0.263	&	0.109	&	0.576 \\
    \fairbase & 0.705	&	0.628	&	0.654	&	0.122	&	0.602	&	0.656	&	0.650	&	0.307	&	0.133	&	0.640 \\
     \mixup & 0.628	&	0.638	&	0.661	&	0.121	&	0.610	&	0.668	&	0.665	&	0.246	&	\textbf{0.099}	&	0.555	 \\
   \mixupeo & 0.703	&	0.633	&	0.659	&	0.120	&	0.606	&	0.661	&	0.658	&	0.298	&	0.132	&	0.636 \\
   \mixupma & 0.713	&	0.608	&	0.635	&	0.126	&	0.587	&	0.639	&	0.630	&	0.303	&	0.131	&	0.649 \\
    \mixupmc & 0.708	&	0.572	&	0.602	&	0.136	&	0.557	&	0.603	&	0.592	&	0.302	&	0.130	&	0.640 \\ 
    \fmdp & 0.715	&	0.642	&	0.667	&	\textbf{0.117}	&	0.614	&	0.669	&	0.667	&	0.307	&	0.128	&	0.655 \\
    \fmeo & 0.700	&	0.638	&	0.662	&	0.119	&	0.610	&	0.666	&	0.663	&	0.295	&	0.131	&	0.630 \\
    \mpma &	0.709	&	0.621	&	0.648	&	0.124	&	0.598	&	0.649	&	0.640	&	0.305	&	0.132	&	0.651\\
    \mpmc &	0.706	&	0.593	&	0.622	&	0.131	&	0.574	&	0.626	&	0.614	&	0.303	&	0.131	&	0.630 \\
     \enforcema &	0.679	&	0.537	&	0.566	&	0.145	&	0.522	&	0.570	&	0.555	&	0.268	&	0.114	&	0.591 \\
    \enforcemc  &	0.559	&	0.482	&	0.530	&	0.161	&	0.486	&	0.522	&	0.504	&	0.250	&	0.147	&	0.478\\
    \mixupenforcemc 	&	\textbf{0.526}	&	\textbf{0.474}	&	\textbf{0.519}	&	0.176	&	\textbf{0.484}	&	\textbf{0.509}	&	\textbf{0.475}	&	\textbf{0.233}	&	0.132	&	\textbf{0.422} \\
    \bottomrule
\end{tabular}
\caption{MC $\alpha$ of least calibrated group for methods across 40 (state, year) pairs $\times$ \employment and \income.}
\label{tab:worst_mc_summary}
\end{table*}
\section{Implementation Details}\label{app:implementation}
\subsection{Neural Network} The NN we used for our experiments was directly taken from \citet{chuang2021fair}. It consists of 3 hidden layers of size 200 using ReLU activation after the first and second layers, and an output layer of size 1 with a sigmoid activation. We use the Adam optimizer with learning rate 0.001, as in \citet{chuang2021fair} as well. We use binary cross-entropy loss and train for 10 epochs with $n = 100$ iterations of mini-batch selection ($b = 500$) each. We use the PyTorch\footnote{\url{https://pytorch.org/}} library. 

\subsection{(Fair) Mixup}
For every mini-batch in the Mixup and Fair Mixup variants, we generate a fresh $t \sim \text{Beta}(\epsilon, \epsilon)$ , with $\epsilon = 1$, as in \citet{chuang2021fair}. Also following \citet{chuang2021fair}, even though the integral goes continuously from 0 to 1 in the mathematical specification of Fair Mixup, we only generate one interpolated point ($t$ of the way from the minority group member to the majority group member of each pair) per pair due to practical time and compute power constraints.

\subsection{Post-Processing Algorithms}\label{app:pp}
To enforce MC, we use $d = 10$, and for both MA and MC enforcement, we use $\alpha = 0.01$ (strictly smaller than the smallest mean MC violation among the results of the post-processing-free methods). We require $\alpha$-calibration to hold on the entirety of $S$ rather than just a $(1 - \alpha)$ fraction of $S$, as enforcing the latter definition would naively require us to calculate the calibration of the predictor w.r.t. each of the $ \binom{|S|}{(1 - \alpha)|S|}$ subsets of $S$ and terminate the algorithm if and only if the predictor was $\alpha$-calibrated on the entirety of at least one of these sets, which would take an unreasonable amount of time.

We treat the predictor as an array-like object of probabilities $\in [0, 1]$ where the $i$th element of the list corresponds to $f_i$. While pseudocode for multicalibrating a predictor is provided in \citet{pmlr-v80-hebert-johnson18a}, it does not account for practical implementation considerations, most importantly the additional output of a list of rules specifying what updates to apply to predictions on future unseen data. Thus, we provide code for the MC algorithm here and in our public repo (to be released upon publication). The inputs to the algorithm are $p$, the initial predictor's outputs; $\mathcal{C}$, the set of subsets of the population that need to be multicalibrated; $y$, a list of true outcomes in $\{0, 1\}$; $\alpha$, the violation parameter, and (for MC only) $d$, the prediction interval discretization parameter.

For prediction adjustments learned from the post-processing set to be useful, they must be applied to a baseline predictor at \textit{test time}. To achieve this, we append the updates made to our post-processing sets to a \texttt{list} and return the list at the end of the algorithms. We then apply the updates in the list one at a time to the predictors at test time. This mimics the conceptualization of the MC algorithm as a circuit, as suggested in \citet{pmlr-v80-hebert-johnson18a}, where a predictor achieves MC on more and more sets as it passes through increasingly large computational gates until it is multicalibrated with respect to the entire superset.

\begin{listing*}[!t]
\caption{Implementation of MC post-processing.}
\label{lst:mc}
\begin{lstlisting}[language=Python]
# Function to calculate MC violations
# Inputs:
    # p: list of predicted probabilities
    # S: list of indices specifying group to consider
    # y: list of true outcomes
    # d: group discretization parameter
# Outputs: 
    # v: index of prediction interval with highest violation
    # highest: value of highest violation at interval v
    # S_alphas: list of violations for all prediction intervals
    # S_vs: list where vth element is the subset of S with predictions in interval v
def calculate_calibration(p, S, y, d):
    # split S into intervals
    S_vs = [[] for _ in range(d + 1)]
    for i in range(d + 1):
        for e in S:
            if e is not None and i / d <= p[e]  < (i + 1) / d:
                S_vs[i].append(l)
    # compute violations               
    S_alphas = [0 for _ in range(d + 1)]
    for interval in range(len(S_vs)):
        n = len(S_vs[interval])
        if n != 0:
            predictor_sum = 0.
            y_sum = 0.
            for e in S_vs[interval]:
                if e is not None:
                    predictor_sum += p[e]; data_sum += y[e]
            S_alphas[interval] = (y_sum - predictor_sum) / n
    # return the violation posed by the interval with the maximum violation
    v_max, v_min = S_alphas.index(max(S_alphas)), S_alphas.index(min(S_alphas))
    v = v_max if abs(S_alphas[v_max]) > abs(S_alphas[v_min]) else v_min  
    highest = abs(S_alphas[v])  
    return v, highest, S_alphas, S_vs

# Function to enforce max MC violation of alpha w.r.t. C on post-processing set X with predictions p (discretized by parameter d) and true outcomes y
# Outputs:
    # p_new: updated predictions on post-processing set
    # updates: list of rules mapping (group, interval) pairs to future prediction adjustments
def multicalibrate(X, p, C, y, alpha, d):
    p_new, done, seen, updates = p.copy(), False, [], []
    while not done:
        # randomly select 1 group at a time to calibrate
        ind = np.random.choice(len(C), size = 1, replace = False)[0]
        A = np.ones(X.shape[0])
        for _, criterion, value in C[ind]:
            A *= X[:, criterion] == value
        S = np.where(A == 1)[0]; seen.append(ind)
        # if X has members of S, calculate and adjust prediction errors
        if len(S) > 0:
            v, highest, S_alphas, S_vs = calculate_calibration(p_new, S, y, d)
            # if violation too high, update p by nudging
            if (highest > alpha):
                seen = []
                updates.append((C[ind], v, S_alphas[v]))
                for el in S_deciles[j]:
                    if e is not None:
                        p_new[e] += S_alphas[v]
                        p_new[e] = min(p_new[e], 1); p_new[e] = max(p_new[e], 0)
        done = len(seen) == len(C)
    return p_new, updates
\end{lstlisting}
\end{listing*}

\begin{listing*}[!t]
\caption{Implementation of MA post-processing.}
\label{lst:ma}
\begin{lstlisting}[language=Python]
# Function to calculate MA violations
# Inputs:
    # p: list of predicted probabilities
    # S: list of indices specifying group to consider
    # y: list of true outcomes
# Output: mean prediction error across elements of S
def calculate_accuracy(p, S, y):
    predictor_sum = 0
    y_sum = 0
    # compute violations           
    for e in S:
        if e is not None:
            predictor_sum += p[e]; y_sum += y[e]
    return (y_sum - predictor_sum) / len(S)

# Function to enforce max MA violation of alpha w.r.t. C on post-processing set X with predictions p and true outcomes y
# Outputs:
    # p_new: updated predictions on post-processing set
    # updates: list of rules mapping groups to future prediction adjustments
def enforce_multiaccuracy(X, p, C, y, alpha):
    p_new, done, seen, updates = p.copy(), False, [], []
    while not done:
        # randomly choose 1 set to start with
        ind = np.random.choice(len(C), size = 1, replace = False)[0]
        A = np.ones(X.shape[0])
         for _, criterion, value in C[ind]:
            A *= X[:, criterion] == value
        S = np.where(A == 1)[0]
        seen.append(ind)
        # if X has members of S, calculate and adjust errors
        if len(S) > 0:
            violation = calculate_accuracy(p_new, S, y)
            # if violation too high, update p by nudging
            if (abs(violation) > alpha):
                seen = []
                updates.append((C[ind], violation))
                for e in S:
                    if e is not None:
                        p_new[e] += violation
                        p_new[e] = min(p_new[e], 1); p_new[e] = max(p_new[e], 0)
        done = len(seen) == len(C):
    return p_new, updates
\end{lstlisting}
\end{listing*}

\section{Full Fair Mixup Specification}\label{app:fm}
In all formulae below, $S' = \neg S$.
\subsection{\fmdp} 
We begin with the metric specification for demographic parity:
\begin{equation}
    \mathcal{M}_S = \Delta \text{DP}_S(f) \\
    = |\mathbb{E}_{x_i \sim S}[f_i] - \mathbb{E}_{x_i \sim S'}[f_i]|.
\end{equation} 
For a continuously differentiable function $T: \mathcal{X}^2 \times [0, 1] \to \mathcal{X}$ such that $T(x_1, x_1', 0) = x_1$ and $T(x_1, x_1', 1) = x_1'$, this metric applied to the interpolated synthetic points would be \begin{equation}
    \Delta \text{DP}_S(f) \\
    =\big \lvert \int_0^1 \frac{d}{dt} \int f(T(x_1, x_1', t) dS(x_1) dS'(x_1') dt\big \rvert.
\end{equation} We refer the reader to \citet{chuang2021fair} for a full proof. The gradient of this metric would be \begin{align}\label{eq:fmdp}
    \mathcal{R}_{\text{mixup}}^{\text{DP}_S} = \int_0^1 \lvert \int \mathcal{I} dS(x_1)dS'(x_1') dt \rvert,
\end{align} where 
\begin{equation}\label{eq:cali}
    \mathcal{I} = \langle \nabla_x f(T(x_1, x_1', t), x_1 - x_1' \rangle.
\end{equation}
\subsection{\fmeo}
Again, we begin with the metric definition: \begin{equation}
    \mathcal{M}_S = \Delta \text{EO}_{S}(f) \\
    = \sum_{y \in \{0, 1\}}|\mathbb{E}_{x_i \sim S_y}[f_i] - \mathbb{E}_{x_i \sim S'_y}[f_i]|,
\end{equation} where $S_y = \{x_i \in S: y_i = y\}$, and $S' = \neg S$. Then, we transform it into an integral to be able to apply it to interpolated points:
\begin{equation}
    \Delta \text{EO}_S(f) =\sum_{y \in \{0, 1\}}\big \lvert \int_0^1 \frac{d}{dt} \int f_T dS_y(x_1) dS'_y(x_1') dt\big \rvert,
\end{equation}
where \begin{equation}
    f_T = f(T(x_1, x_1', t).
\end{equation}
Next, we take the Jacobian to get our per-group penalty: \begin{align}\label{eq:fmeo}
    \mathcal{R}_{\text{mixup}}^{\text{EO}_{S}} = \sum_{y \in \{0, 1\}} \int_0^1 \Bigg \lvert \int \mathcal{I} dS_y(x_1)dS'_y(x_1') dt\Bigg \rvert,
\end{align} with $\mathcal{I}$ defined as in Equation \ref{eq:cali}.
\subsection{\mpma}
We begin with our metric specification for MA:
\begin{equation}
    \mathcal{M}_S = \Delta \text{MA}_S(f) \\
    = |\mathbb{E}_{x_i \sim S}[e_i] - \mathbb{E}_{x_i \sim S'}[e_i]|,
\end{equation} 
Transforming this into an integral, we obtain:
\begin{equation}
    \Delta{\text{MA}}_S(f) = \big \lvert \int_0^1 \frac{d}{dt} (\mu_f(t) - \mathbb{E}_{(x_1, x_1') \sim (S, S')}[y(t)]) dt \big \rvert,
\end{equation} where
\begin{equation}
    \mu_f(t) = \mathbb{E}_{(x_1, x_1') \sim (S, S')}[f(tx_1 + (1 - t)x_1')],
\end{equation} and \begin{equation}\label{eq:yt}
    y(t) = ty_1 + (1 - t)y_1'.
\end{equation}
Taking the Jacobian, we have \begin{equation}
    R_{\text{mixup}}^{\text{MA}_S} \\
    = \int_0^1 \Bigg\lvert  \int \langle\nabla_x \mathcal{T}, x_1 - x_1' \rangle dS(x_1)d S'(x_1')\Bigg\rvert dt,
\end{equation} where 
\begin{align}\label{eq:t}
    \mathcal{T} &= f(tx_1 + (1 - t)x_1') - (ty_1 + (1 - t) y_1').
\end{align}
\subsection{\mpmc}
Beginning with our metric definition: \begin{equation}
    \mathcal{M}_S = \Delta \text{MC}_S(f) = \sum_{v = 0}^d |\mathbb{E}_{x_i \sim S_{v}^d}[e_i] - \mathbb{E}_{x_i \sim S'^d_{v}}[e_i]|,
\end{equation} 
where $S_{v}^d = \{ x_i \in S: f_i \in [\frac{v}{d}, \frac{v + 1}{d}) \}$,
we create our integral as follows:
\begin{equation}
   \Delta{\text{MC}}_{S_{v}^d} = \big \lvert \int_0^1 \frac{d}{dt} (\mu_f(t) - \mathbb{E}_{(x_1, x_1') \sim (S_v^d, S'^d_v)}[y(t)]) dt \big \rvert,
\end{equation}
where
\begin{equation}
    \mu_f(t) = \mathbb{E}_{x_1, x_1' \sim S_v^d, S'^d_v}[f(tx_1 + (1 - t)x_1')],
\end{equation} with $y(t)$ as in Eq. \ref{eq:yt}. Then, \begin{equation}
    \mathcal{R}_{\text{mixup}}^{\text{MC}_{S_v^d}} \\
    = \int_0^1 \Bigg\lvert  \int \langle\nabla_x \mathcal{T}, x_1 - x_1' \rangle dS_{v}^d(x_1)d S'^d_v(x_1')\Bigg\rvert dt,
\end{equation}
with $\mathcal{T}$ as in Eq. \ref{eq:t}. Finally, we sum over the prediction intervals to get \begin{equation}
    \mathcal{R}_{\text{mixup}}^{\text{MC}_S} = \sum_{v = 0}^d \mathcal{R}_{\text{mixup}}^{\text{MC}_{S_v^d}}.
\end{equation}

\bibliography{aaai25}

\end{document}